\documentclass{article} %
\usepackage{times}
\usepackage[accepted]{icml2021}
\usepackage{natbib}
\usepackage{microtype}
\usepackage{float}
\usepackage{stfloats}
\usepackage[pdftex]{graphicx} %
\usepackage{subfigure}
\usepackage{booktabs}
\usepackage{makecell}
\usepackage[utf8]{inputenc} %
\usepackage[T1]{fontenc}    %
\usepackage{hyperref} %
\usepackage{url}            %
\usepackage{amsfonts}       %
\usepackage{nicefrac}       %
\usepackage{microtype}      %
\usepackage[font=small]{caption} %
\usepackage{amsmath}        %
\usepackage{amsthm}         %
\usepackage{wrapfig}        %
\usepackage[ruled]{algorithm2e}
\usepackage{listings}       %
\usepackage{bm}             %
\usepackage{color}
\usepackage{tikz}
\usepackage{tabu}
\usepackage{makecell}
\usepackage{multicol}

\usepackage{amsmath,amsfonts,bm}

\def\Figref#1{Figure~\ref{#1}}

\def\Tabref#1{Table~\ref{#1}}

\def\Secref#1{Section~\ref{#1}}

\def\eqref#1{equation~(\ref{#1})}
\def\Eqref#1{Equation~(\ref{#1})}

\def\Algref#1{Algorithm~\ref{#1}}

\def\1{\bm{1}}

\DeclareMathAlphabet{\mathsfit}{\encodingdefault}{\sfdefault}{m}{sl}
\SetMathAlphabet{\mathsfit}{bold}{\encodingdefault}{\sfdefault}{bx}{n}

\newcommand{\E}{\mathbb{E}}

\newcommand{\KL}{D_{\mathrm{KL}}}

\DeclareMathOperator*{\argmax}{arg\,max}

\usepackage{colortbl}
\usepackage{bbm}
\usepackage{multirow}
\usepackage[export]{adjustbox}
\usepackage[normalem]{ulem}
\usepackage{cleveref}
\usepackage{bbm}
\usepackage{multirow}
\usepackage[export]{adjustbox}
\usepackage[normalem]{ulem}
\usepackage{cleveref}
\usepackage{threeparttable}
\usepackage{lipsum}
\usepackage{amssymb}
\usepackage{pifont}
\usepackage[makeroom]{cancel}
\newcommand{\cmark}{\ding{51}}
\newcommand{\xmark}{\ding{55}}
\definecolor{Gray}{gray}{0.5}
\definecolor{GrayBG}{gray}{0.95}
\usepackage{etoolbox}

\makeatletter
\AfterEndEnvironment{algorithm}{\let\@algcomment\relax}
\AtEndEnvironment{algorithm}{\kern2pt\hrule\relax\vskip3pt\@algcomment}
\let\@algcomment\relax
\newcommand\algcomment[1]{\def\@algcomment{\footnotesize#1}}
\renewcommand\fs@ruled{\def\@fs@cfont{\bfseries}\let\@fs@capt\floatc@ruled
  \def\@fs@pre{\hrule height.8pt depth0pt \kern2pt}%
  \def\@fs@post{}%
  \def\@fs@mid{\kern2pt\hrule\kern2pt}%
  \let\@fs@iftopcapt\iftrue}
\makeatother

\newcommand{\titlename}{APS: Active Pretraining with Successor Features}
\newcommand{\shorttitle}{{Active Pretraining with Successor Features}}
\newcommand{\ours}{{APS}}

\begin{document}

\icmltitlerunning{\shorttitle{}}
\twocolumn[
\icmltitle{\titlename{}}

\begin{icmlauthorlist}
\icmlauthor{Hao Liu}{bk}
\icmlauthor{Pieter Abbeel}{bk}
\end{icmlauthorlist}

\icmlaffiliation{bk}{University of California, Berkeley, CA, USA}

\icmlcorrespondingauthor{Hao Liu}{hao.liu@cs.berkeley.edu}

\vskip 0.3in
]

\printAffiliationsAndNotice{}  %

\begin{abstract}
We introduce a new unsupervised pretraining objective for reinforcement learning. During the unsupervised reward-free pretraining phase, the agent maximizes mutual information between tasks and states induced by the policy. Our key contribution is a novel lower bound of this intractable quantity. We show that by reinterpreting and combining variational successor features~\citep{Hansen2020Fast} with nonparametric entropy maximization~\citep{liu2021behavior}, the intractable mutual information can be efficiently optimized. The proposed method Active Pretraining with Successor Feature (APS) explores the environment via nonparametric entropy maximization, and the explored data can be efficiently leveraged to learn behavior by variational successor features. APS addresses the limitations of existing mutual information maximization based and entropy maximization based unsupervised RL, and combines the best of both worlds. When evaluated on the Atari 100k data-efficiency benchmark, our approach significantly outperforms previous methods combining unsupervised pretraining with task-specific finetuning.

\end{abstract}

\section{Introduction}

\begin{figure*}[t]
\centering
\includegraphics[width=.90\textwidth]{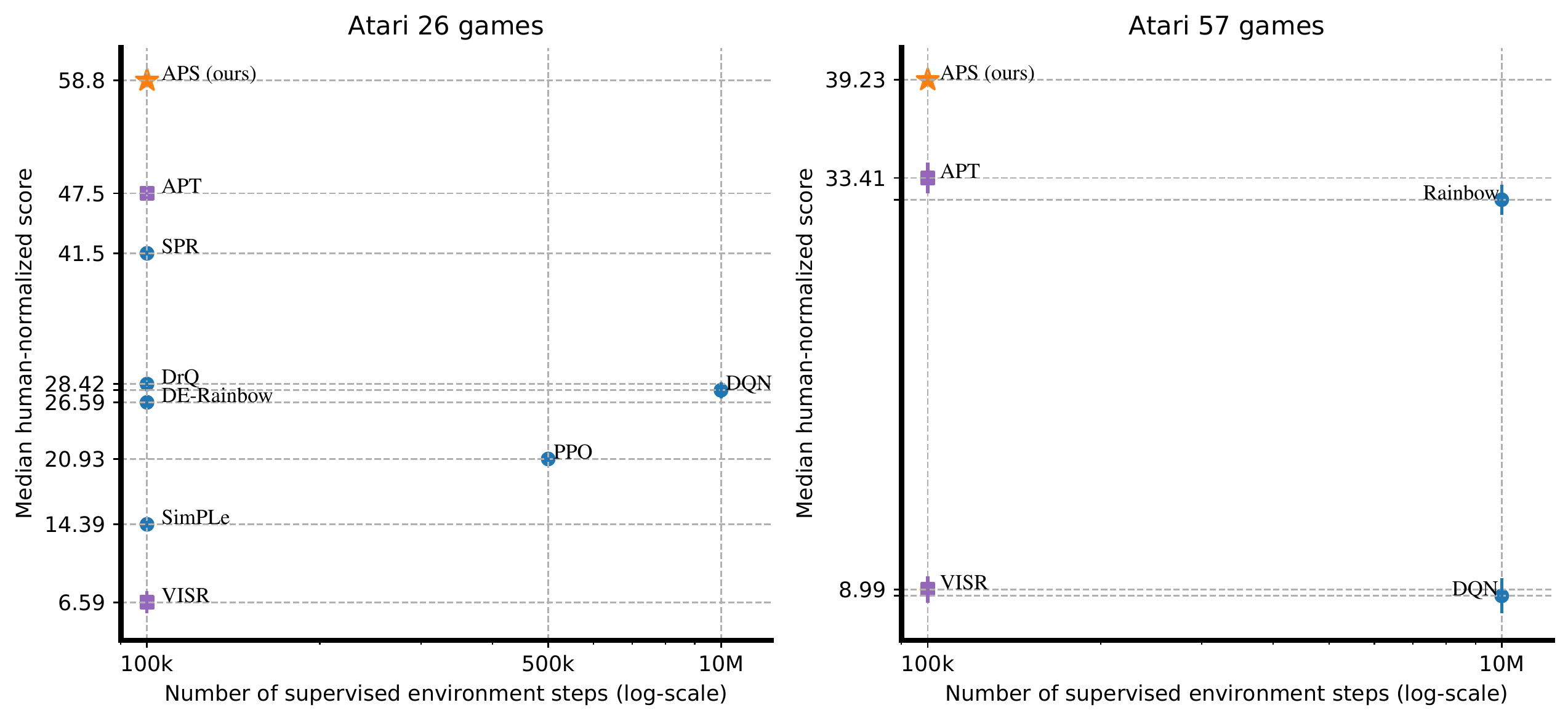}
\caption{
Median of human normalized score on the 26 Atari games considered by~\citet{kaiser2019model} (left) and the Atari 57 games considered in~\citet{mnih2015human}(right).
Fully supervised RL baselines are shown in circle.
RL with unsupervised pretraining are shown in square.
\ours{} significantly outperforms all of the fully supervised and unsupervised pre-trained RL methods.
Baselines: Rainbow~\citep{hessel2018rainbow}, SimPLe~\citep{kaiser2019model}, APT~\citep{liu2021behavior}, Data-efficient Rainbow~\citep{kielak2020recent}, DrQ~\citep{kostrikov2020image}, VISR~\citep{Hansen2020Fast}, CURL~\citep{srinivas2020curl},  and SPR~\citep{schwarzer2021dataefficient}. 
}
\label{fig:atari_imagenet}
\end{figure*}

Deep unsupervised pretraining has achieved remarkable success in various frontier AI domains from natural language processing~\citep{devlin2018bert, peters2018deep, brown2020language} to computer vision~\citep{he2019momentum, chen2020simple}.
The pre-trained models can quickly solve downstream tasks through few-shot fine-tuning~\citep{brown2020language,chen2020big}.

In reinforcement learning (RL), however, training from scratch to maximize extrinsic reward is still the dominant paradigm.
Despite RL having made significant progress in 
playing video games~\citep{mnih2015human, schrittwieser2019mastering, vinyals2019grandmaster, badia2020agent57} and solving complex robotic control tasks~\citep{andrychowicz2017hindsight, akkaya2019solving}, RL algorithms have to be trained from scratch to maximize extrinsic return for every encountered task.
This is in sharp contrast with how intelligent creatures quickly adapt to new tasks by leveraging previously acquired behaviors. 

In order to bridge this gap, unsupervised pretraining RL has gained interest recently, from state-based~\citep{gregor2016variational, eysenbach2018diversity, sharma2019dynamics, mutti2020policy} to pixel-based RL~\citep{Hansen2020Fast, liu2021behavior, campos2021coverage}.
In unsupervised pretraining RL, 
the agent is allowed to train for a long period without access to environment reward, and then got exposed to reward during testing. 
The goal of pretraining is to have data efficient adaptation for some downstream task defined in the form of rewards.

State-of-the-art unsupervised RL methods consider various ways of designing the so called intrinsic reward~\citep{barto2004intrinsically, barto2013intrinsic, gregor2016variational, achiam2017surprise}, with the goal that maximizing this intrinsic return can encourage meaningful behavior in the absence of external rewards.
There are two lines of work in this direction, we will discuss their advantages and limitations, and show that a novel combination yields an effective algorithm which brings the best of both world.

The first category is based on maximizing the mutual information between task variables ($p(z)$) and their behavior in terms of state visitation ($p(s)$) to encourage learning distinguishable task conditioned behaviors, which has been shown effective in state-based RL~\citep{gregor2016variational,eysenbach2018diversity} and visual RL~\citep{Hansen2020Fast}.
VISR proposed in~\citet{Hansen2020Fast} is the prior state-of-the-art in this category. 
The objective of VISR is $\max I(s; z) = \max H(z) - H(s|z)$ where $z$ is sampled from a fixed distribution. 
VISR proposes a successor features based variational approximation to maximize a variational lower bound of the intractable conditional entropy $-H(s|z)$. 
The advantage of VISR is that its successor features can quickly adapt to new tasks.
Despite its effectiveness, the fundamental problem faced by VISR is lack of exploration.

Another category is based on maximizing the entropy of the states induced by the policy $\max H(s)$.
Maximizing state entropy has been shown to work well in state-based domains~\citep{hazan2019provably, mutti2020policy} and pixel-based domains~\citep{liu2021behavior}.
It is also shown to be provably efficient under certain assumptions~\citep{hazan2019provably}.
The prior state-of-the-art APT by~\citet{liu2021behavior} show maximizing a particle-based entropy in a lower dimensional abstraction space can boost data efficiency and asymptotic performance.
However, the issues with APT are that it is purely exploratory and task-agnostic and lacks of the notion of task variables, making it more difficult to adapt to new tasks compared with task-conditioning policies.

Our main contribution is to address the issues of APT and VISR by combining them together in a novel way.
To do so, we consider the alternative direction of maximizing mutual information between states and task variables $I(s;z)=H(s)-H(s|z)$,
the state entropy $H(s)$ encourages exploration while the conditional entropy encourages the agent to learn task conditioned behaviors.
Prior work that considered this objective had to either make the strong assumption that the distribution over states can be approximated with the stationary state-distribution of the
policy~\citep{sharma2019dynamics} or rely on the challenging density modeling to derive a tractable lower bound~\citep{sharma2019dynamics, campos2020explore}. %
We show that 
the intractable conditional entropy, $-H(s|z)$ can be lower bounded and optimized by learning successor features. 
We use APT to maximize the state entropy $H(s)$ in an abstract representation space.
Building upon this insight, we propose Active Pretraining with Successor Features (\ours{}) since the agent is encouraged to actively explore and leverage the experience to learn behavior. 
By doing so, we experimentally find that they address the limitations of each other and significantly improve each other.

We evaluate our approach on the Atari benchmark~\citep{bellemare2013arcade} where we apply \ours{} to DrQ~\citep{kostrikov2020image} and test its performance after fine-tuning for 100K supervised environment steps.
The results are shown in~\Figref{fig:atari_imagenet}.
On the 26 Atari games considered by~\citep{kaiser2019model}, our fine-tuning signiﬁcantly boosts the data-efﬁciency of DrQ, achieving 106\% relative improvement. 
On the full suite of Atari 57 games~\citep{mnih2015human}, fine-tuning \ours{} pre-trained models significantly outperforms prior state-of-the-art, achieving human median score $3\times$ higher than DQN trained with 10M supervised environment steps and outperforms previous methods combining unsupervised pretraining with task-specific finetuning.

\section{Related Work}
Our work falls under the category of mutual information maximization for unsupervised behavior learning.
Unsupervised discovering of a set of task-agnostic behaviors by means of seeking to maximize an extrinsic reward has been explored in the the evolutionary computation community~\citep{lehman2011abandoning, lehman2011novelty}. 
This has long been studied as intrinsic motivation~\citep{barto2013intrinsic, barto2004intrinsically}, often with the goal of encouraging exploration~\citep{csimcsek2006intrinsic, oudeyer2009intrinsic}.
Entropy maximization in state space has been used to encourage exploration in state RL~\citep{hazan2019provably, mutti2020policy, seo2021state} and visual RL~\citep{liu2021behavior, yarats2021reinforcement}.
Maximizing the mutual information between latent variable policies and their behavior in terms of state visitation has been used as an objective for discovering meaningful behaviors~\citep{houthooft2016curiosity, mohamed2015variational, gregor2016variational, houthooft2016vime, eysenbach2018diversity, warde2018unsupervised}.
\citet{sharma2019dynamics} consider a similar decomposition of mutual information, namely, $I(s;z) = H(s) - H(z|s)$, however, they assume $p(s|z) \approx p(s)$ to derive a different lower-bound of the marginal entropy. 
Different from~\citet{sharma2019dynamics},
~\citet{campos2020explore} propose to first maximize $H(s)$ via maximum entropy estimation~\citep{hazan2019provably, lee2019efficient} then learn behaviors, this method relies on a density model that provides an estimate of how many times an action has been taken in similar states.
These methods are also only shown to work from explicit state-representations, and it is nonobvious how to modify them to work from pixels.
The work by~\citet{badia2020never} also considers k-nearest neighbor based count bonus
to encourage exploration, yielding improved performance on Atari games.
This heuristically defined count-based bonus has been shown to be an effective unsupervised pretraining objective for RL~\citep{campos2021coverage}.
~\citet{machado2018count} show the norm of learned successor features can be used to incentivize exploration as a reward bonus.
Our work differs in that we jointly maximize the entropy and learn successor features.

\begin{table*}
\centering
\caption{Comparing methods for pretraining RL in no reward setting. VISR~\citep{Hansen2020Fast}, APT~\citep{liu2021behavior}, MEPOL~\citep{mutti2020policy}, DIYAN~\citep{eysenbach2018diversity}, DADS~\citep{sharma2019dynamics}, EDL~\citep{campos2020explore}.
Exploration: the model can explore efficiently.
Off-policy: the model is off-policy RL.
Visual: the method works well in visual RL, e.g., Atari games.
Task: the model conditions on latent task variables $z$. 
$^\star$ means only in state-based RL.
}
\label{tab:model_summary}
\begin{tabular}{lllllll}
\toprule
Algorithm & Objective & Exploration & Visual & Task & Off-policy & Pre-Trained Model \\
\midrule 
APT & $\max \mathrm{H}(s)$ & \cmark &  \cmark & \xmark & \cmark & $\pi(a|s), Q(s, a)$  \\
VISR & $\max \mathrm{H}(z) - \mathrm{H}(z|s)$ & \xmark &  \cmark & \cmark  & \cmark & $\psi(s,z)$, $\phi(s)$  \\
\midrule
MEPOL & $\max \mathrm{H}(s)$ & \cmark$^\star$ & \xmark & \xmark & \xmark &  $\pi(a|s)$ \\
DIAYN & $\max - \mathrm{H}(z|s) + \mathrm{H}(a|z, s)$ & \xmark  & \xmark & \cmark & \xmark & $\pi(a|s,z)$ \\
EDL & $\max \mathrm{H}(s) - \mathrm{H}(s|z)$ & \cmark$^\star$ &  \xmark & \cmark & \cmark & $\pi(a|s, z), q(s'|s, z)$ \\
DADS & $\max \mathrm{H}(s) - \mathrm{H}(s|z)$ & \cmark & \xmark & \cmark & \xmark & $\pi(a|s, z), q(s'|s, z)$ \\
\midrule
\ours{} & $\max \mathrm{H}(s) - \mathrm{H}(s|z)$ & \cmark & \cmark & \cmark & \cmark & $\psi(s,z)$, $\phi(s)$ \\
\bottomrule
\multicolumn{7}{l}{
\begin{small} 
$\psi(s)$: successor features, $\phi(s)$: state feature ($i.e.$, the representation of states). 
\end{small}}
\\
\end{tabular}
\vskip -0.2in
\end{table*}

\section{Preliminaries}
Reinforcement learning considers the problem of finding an optimal policy for an agent that interacts with an uncertain environment and collects reward per action.
The goal of the agent is to maximize its cumulative reward. 

Formally, this problem can be viewed as a Markov decision process (MDP) defined by $(\mathcal{S}, \mathcal{A}, \mathcal{T}, \rho_0, r, \gamma)$ where
$\mathcal{S} \subseteq \mathbb{R}^{n_s}$ is a set of $n_s$-dimensional states,
$\mathcal{A} \subseteq \mathbb{R}^{n_a}$ is a set of $n_a$-dimensional actions,
$\mathcal{T}: \mathcal{S} \times \mathcal{A} \times \mathcal{S} \to [0, 1]$
is the state transition probability distribution. 
$\rho_0: \mathcal{S} \to [0, 1]$ is the distribution over initial states,
$r: \mathcal{S} \times \mathcal{A} \to \mathbb{R}$ is the reward function, and
$\gamma \in [0, 1)$ is the discount factor.
At environment states $s \in {\it \mathcal{S}}$, the agent take actions $a \in {\it \mathcal{A}}$, in the (unknown) environment dynamics defined by the transition probability $T(s'|s,a)$, and the reward function yields a reward immediately following the action $a_t$ performed in state $s_t$.
We deﬁne the discounted return $G(s_t, a_t) = \sum_{l=0}^\infty \gamma^l r(s_{t+l}, a_{t+l})$ as the discounted sum of future rewards collected by the agent.
In value-based reinforcement learning, the agent learns learns an estimate of the expected discounted return, a.k.a, state-action value function.
\begin{align*}
    Q^\pi(s, a) = \mathbb{E}_{\substack{s_{t}=s \\a_{t}=a}}\left[\sum_{l=0}^\infty \gamma^l r(s_{t+l}, a_{t+l}, s_{t+l+1})\right].
\end{align*}

\subsection{Successor Features}
Successor features~\citep{dayan1993improving, kulkarni2016deep, barreto2017successor, barreto2018transfer} assume that there exist features $\phi(s, a, s') \in R^d$ such that the reward function which speciﬁes a task of interest can be written as
\begin{align*}
    r(s, a, s') = \phi(s, a, s')^T w,
\end{align*}
where $w \in R^d$ is the task vector that specify how desirable each feature component is.

The key observation is that the state-action value function can be decomposed as a linear form~\citep{barreto2017successor}
\begin{align*}
    Q^{\pi}(s, a)
    & =\mathbb{E}_{\substack{s_{t}=s\\ a_{t}=a}}\left[\sum_{i=t}^{\infty} \gamma^{i-t} \phi(s_{i+1}, a_{i+1}, s'_{i+1}) \right]^T w \\
    & \equiv \psi^{\pi}(s, a)^T w,
\end{align*}
where $\psi^{\pi}(s, a)$ are the successor features of $\pi$. 
Intuitively, $\psi(s, a)$ can be seen as a generalization of $Q(s, a)$ to multidimensional value function with reward $\phi(s, a, s')$

\begin{figure*}[t]
\centering
\includegraphics[width=.75\textwidth]{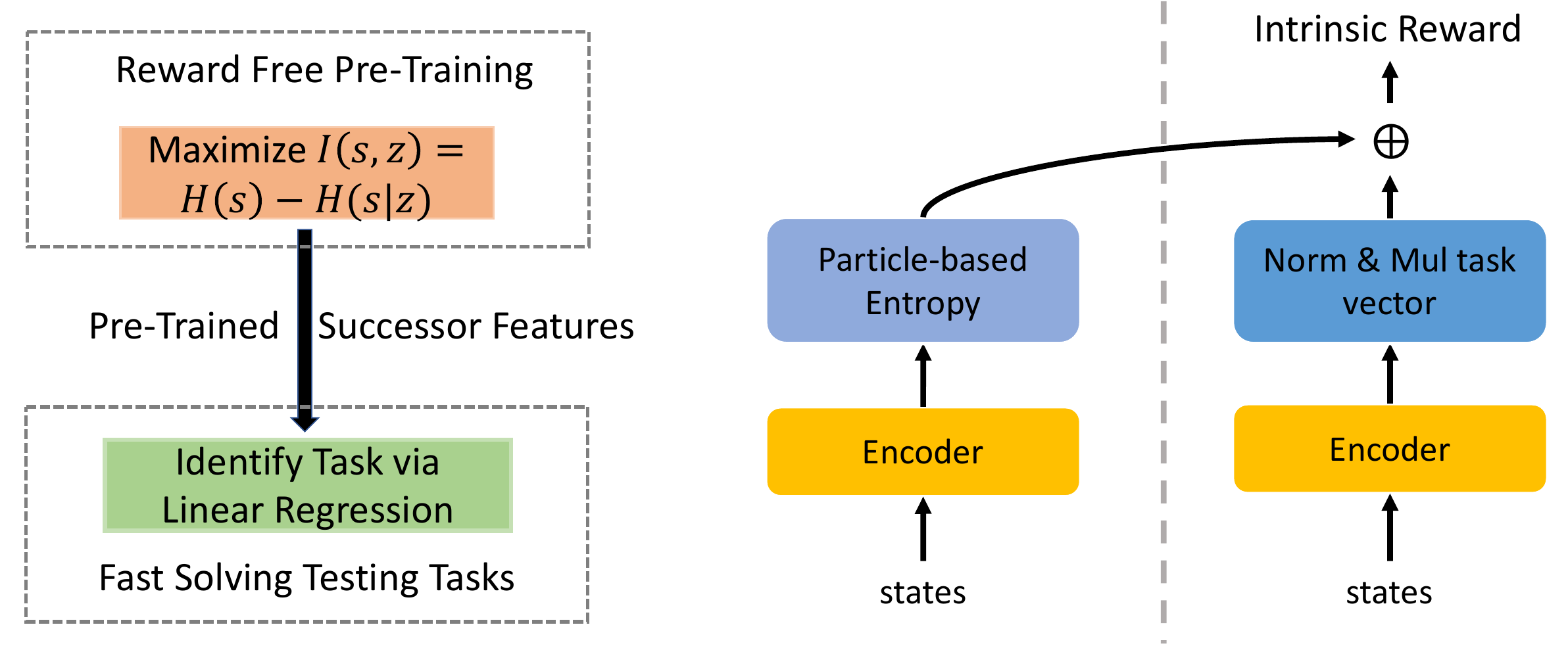}
\caption{
Diagram of the proposed method \ours{}.
On the left shows the concept of \ours{}, during reward-free pretraining phase, reinforcement learning is deployed to maximize the mutual information between the states induced by policy and the task variables.
During testing, the pre-trained state features can identify the downstream task by solving a linear regression problem , the pre-trained task conditioning successor features can then quickly adapt to and solve the task.
On the right shows the components of \ours{}. 
\ours{} consists of maximizing state entropy in an abstract representation space (exploration, $\max H(s)$) and leveraging explored data to learn task conditioning behaviors (exploitation, $\max -H(s|z)$).
}
\label{fig:model_diagram}
\end{figure*}

\section{Method}
We first introduce two techniques which our method builds upon in~\Secref{sec:visr} and~\Secref{sec:apt} and discuss their limitations.
We provide preliminary evidence of the limitations in~\Secref{sec:limitation}. 
Then we propose \ours{} in~\Secref{sec:aps} to address their limitations.

\subsection{Variational Intrinsic Successor Features (VISR)}
\label{sec:visr}

The variational intrinsic successor features (VISR) maximizes the mutual information($I$) between some policy-conditioning variable ($z$) and the states induced by the conditioned policy,
\begin{align*}
    I(z; s) = H(z) - H(z|s),
\end{align*}
where it is common to assume $z$ is drawn from a ﬁxed distribution for the purposes of training stability~\citep{eysenbach2018diversity,Hansen2020Fast}.

This simpliﬁes the objective to minimizing the conditional entropy of the conditioning variable, where $s$ is sampled uniformly over the trajectories induced by $\pi_\theta$.
\begin{align*}
    \sum_{z, s} p(s, z) \log p(z|s) = \E_{s, z}[\log p(z|s)],
\end{align*}
A variational lower bound is proposed to
address the intractable objective,
\begin{align*}
    J_{\text{VISR}}(\theta) = - \E_{s, z}[\log q(z|s)],
\end{align*}
where $q(z|s)$ is a variational approximation.
REINFORCE algorithm is used to learn the policy parameters by treating $\log q(z|s)$ as intrinsic reward.
The variational parameters can be optimized by maximizing log likelihood of samples.

\begin{figure}[b]
\centering
\setlength{\tabcolsep}{1pt} 
\renewcommand{\arraystretch}{1}
\begin{tabular}{cc}
    \includegraphics[width=.22\textwidth]{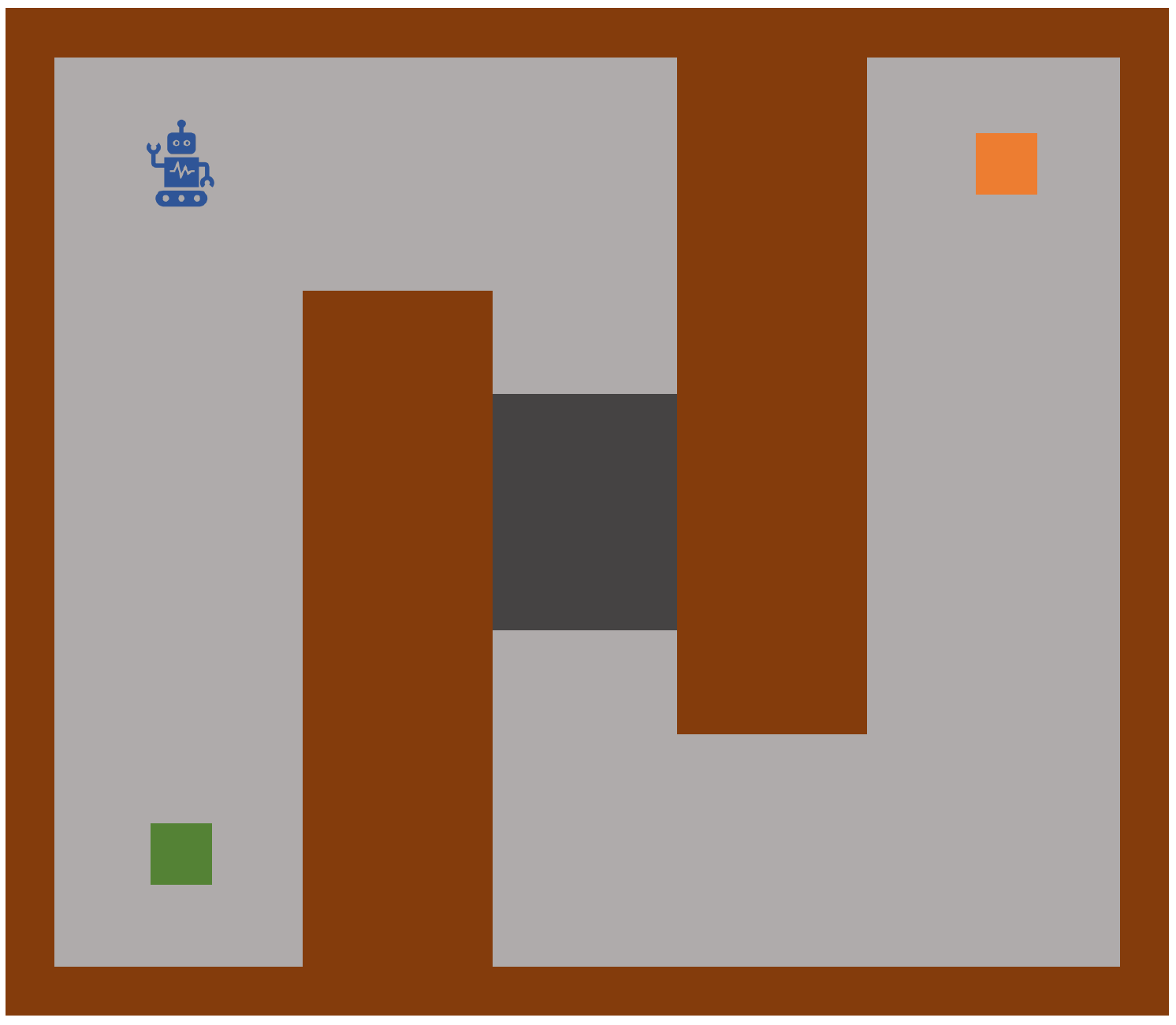} & \includegraphics[width=.22\textwidth]{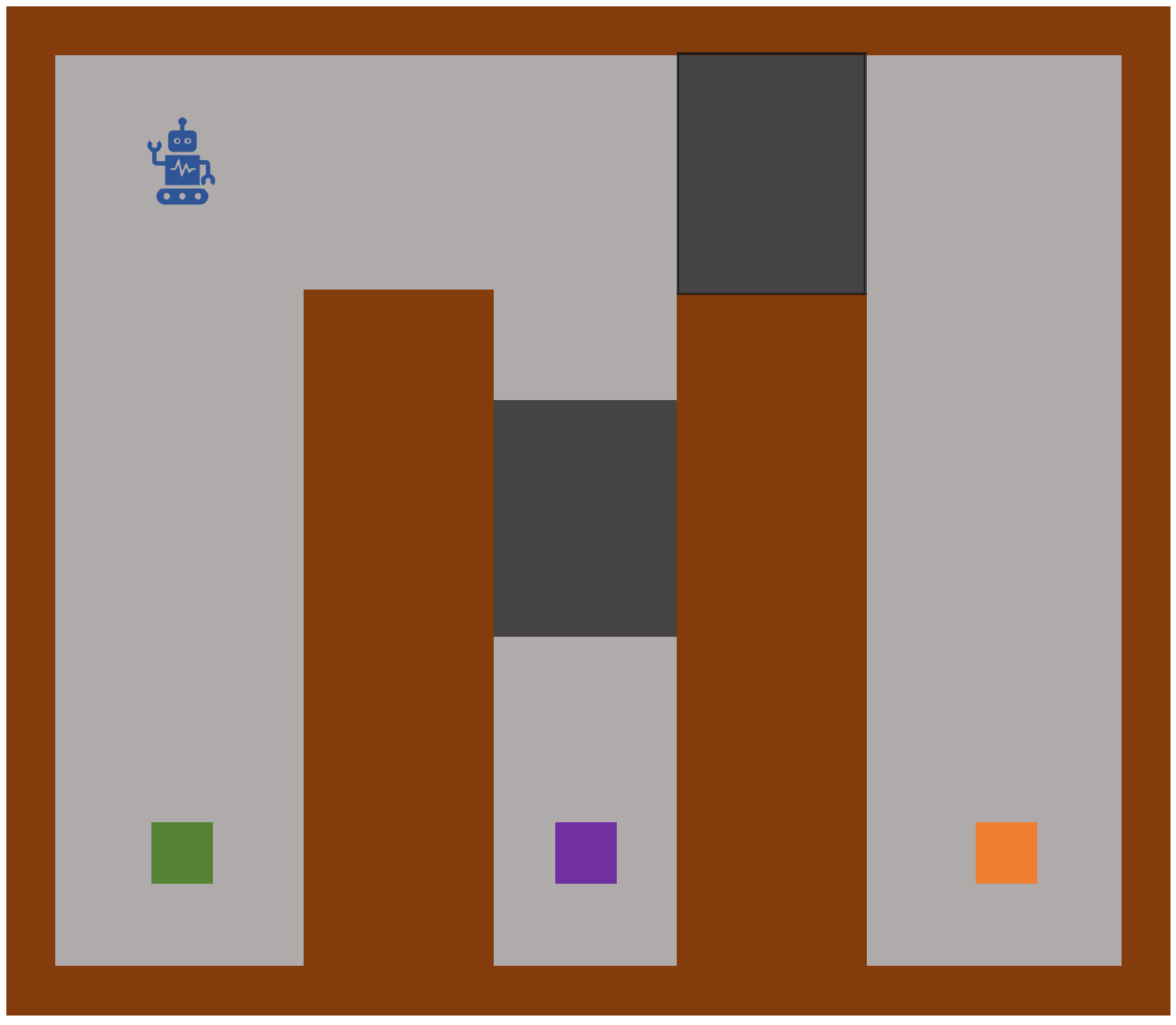} \\
\end{tabular}
\caption{The passageway gridworld environments used in our experiments.
On the left, the agent needs to fetch the key first by navigating to the green location to unlock the closed passageway (shown in black). Similarly, on the right, there is an additional key-passageway pair. The agent must fetch the key (shown in purple) to unlock the upper right passageway.}
\label{fig:gridworld_env}
\end{figure}
The key observation made by~\citet{Hansen2020Fast} is restricting conditioning vectors $z$ to correspond to task-vectors $w$ of the successor features formulation $z \equiv w$.
To satisfy this requirement, one can restrict the task vectors $w$ and features $\phi(s)$ to be unit length and paremeterizing the discriminator $q(z|s)$ as the Von Mises-Fisher distribution with a scale parameter of 1.
\begin{align*}
    r_{\text{VISR}}(s, a, s') = \log q(w|s) = \phi(s)^T w.
\end{align*}
VISR has the rapid task inference mechanism provided by successor features with the ability of mutual information maximization methods to learn many diverse behaviors in an unsupervised way.
Despite its effectiveness as demonstrated in~\citet{Hansen2020Fast}, VISR suffers from inefficient exploration. This issue limits the further applications of VISR in challenging tasks.

\subsection{Unsupervised Active Pretraining (APT)}
\label{sec:apt}
The objective of unsupervised active pretraining (APT) is to maximize the entropy of the states induced by the policy, which is computed in a lower dimensional abstract representation space.
\begin{align*}
    J_{\text{APT}}(\theta) = H(h) = \sum_{s} p(h) \log p(h), \quad h=f(s),
\end{align*}
where $f: R^{n_s} \rightarrow R^{n_h}$ is a mapping that maps observations $s$ to lower dimensional representations $h$.
In their work,~\citet{liu2021behavior} learns the encoder by contrastive representation learning.

With the learned representation, APT shows the entropy of $h$ can be approximated by a particle-based entropy estimation~\citep{singh2003nearest, beirlant1997nonparametric}, which is based on the distance between each particle $h_i = f(s_i)$ and its $k$-th nearest neighbor $h_{i}^{\star}$.
\begin{align*}
H(h) \approx H_{\text{APT}}(h) \propto \sum_{i=1}^{n} \log \|h_{i}-h_{i}^{\star}\|_{n_z}^{n_z}.
\end{align*}
This estimator is asymptotically unbiased and consistent $ \lim_{n \rightarrow \infty} H_{\text{APT}}(s) = H(s)$.

It helps stabilizing training and improving convergence in practice to average over all $k$ nearest neighbors~\citep{liu2021behavior}.
\begin{align*}
\hat{H}_{\text{APT}}(h) = \sum_{i=1}^{n} \log \left(1 + \frac{1}{k} \sum_{h_{i}^{j} \in \mathrm{N}_k(h_i)} \|h_{i}-h_{i}^{j}\|_{n_h}^{n_h} \right),
\end{align*}
where $\mathrm{N}_k(\cdot)$ denotes the $k$ nearest neighbors.

For a batch of transitions $\{(s, a, s')\}$ sampled from the replay buffer, each abstract representation $f(s')$ is treated as a particle and we associate each transition with a intrinsic reward given by
\begin{align}
r_{\text{APT}}(s, a, s') & = \log \left(1 + \frac{1}{k} \sum_{h^{(j)} \in \mathrm{N}_k(h)} \|h-h^{(j)}\|_{n_z}^{n_z} \right) \nonumber \\
\text{where}~h & = f_\theta(s').
\label{eq:re_reward}
\end{align}

While APT achieves prior state-of-the-art performance in DeepMind control suite and Atari games, it does not conditions on latent variables (e.g. task) to capture important task information during pretraining, making it inefficient to quickly identity downstream task when exposed to task specific reward function. 

\begin{figure}[!htbp]
\centering
\includegraphics[width=.48\textwidth]{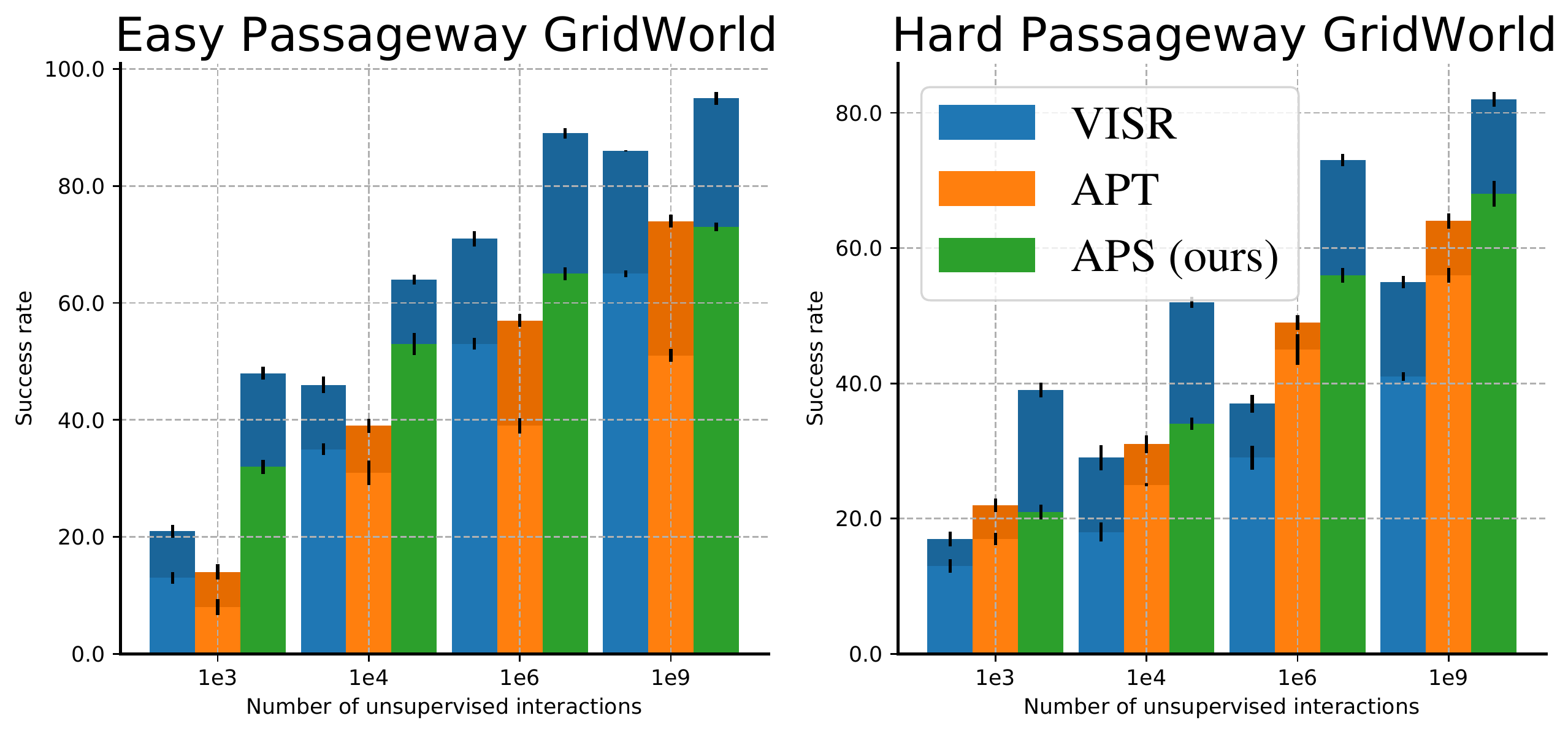}
\caption{Performance of different methods on the gridworld environments in~\Figref{fig:gridworld_env}. 
The results are recorded during testing phase after pretraining for a number of unsupervised interactions.
The success rate are aggregated over 10 random seeds.
The bottom of each bar is the zero-shot testing performance while the top is the fine-tuned performance.
}
\label{fig:gridworld_result}
\end{figure}

\subsection{Empirical Evidence of the Limitations of Existing Models}
\label{sec:limitation}
In this section we present two multi-step grid-world environments to illustrate the drawbacks of APT and VISR, and highlight the importance of both exploration and task inference.
The environments, implemented with the pycolab game engine~\citep{stepleton2017pycolab}, are depicted shown in~\Figref{fig:gridworld_env}, and are fully observable to the agent.
At each episode, the agent starts from a randomly initialized location in the top left corner, with the task of navigating to the target location shown in orange.
To do so, the agent has to first pick up a key(green, purple area) that opens the closed passageway. 
The easy task shown in left of~\Figref{fig:gridworld_env} has one key and one corresponding passageway while the hard task has two key-passageway pairs.
We evaluate the agent in terms of success rates.
During evaluation, the agent receives an intermediate reward 1 for picking up key and 10 for completing the task.
The hierarchical task presents a challenge to algorithms using only exploration bonus or successor features, as the exploratory policy is unlikely to quickly adapt to the task specific reward and the successor features is likely to never explore the space sufficiently. 

\Figref{fig:gridworld_result} shows the success rate of each method.
APT performs worse than VISR at the easy level, possibly because successor features can quickly adapt to the downstream reward.
On the other hand, APT significantly outperforms VISR at the hard level which requires a exploratory policy. 
Despite the simplicity, these two gridworld environments already highlight the weakness of each method.
This observation conﬁrms that existing formulations either fail due to inefficient exploration or slow adaption, and motivates our study of alternative methods for behavior discovery.

\subsection{Active Pre-training with Successor Features}
\label{sec:aps}

To address the issues of APT and VISR, we consider maximizing the mutual information between task variable ($z$) drawn from a fixed distribution and the states induced by the conditioned policy.
\begin{align*}
    I(z;s) = H(s) - H(s|z).
\end{align*}

The intuition is that the $H(s)$ encourages the agent to explore novel states while $H(s|z)$ 
encourages the agent to leverage the collected data to capture task information.

Directly optimizing $H(s)$ is intractable because the true distribution of state is unknown, as introduced in~\Secref{sec:apt}, APT~\citep{liu2021behavior} is an effective approach for maximizing $H(s)$ in high-dimensional state space. We use APT to perform entropy maximization.
\begin{align}
r_{\text{APS}}^{\text{exploration}}(s, a, s') & = \log \left(1 + \frac{1}{k} \sum_{h^{(j)} \in \mathrm{N}_k(h)} \|h-h^{(j)}\|_{n_h}^{n_h} \right) \nonumber \\
\text{where}~h & = f_\theta(s').
\label{eq:aps_exploration}
\end{align}

As introduced in~\Secref{sec:visr}, VISR~\citep{Hansen2020Fast} is a variational based approach for maximizing $-H(z|s)$.
However, maximizing $-H(z|s)$ is not directly applicable to our case where the goal is to maximize $-H(s|z)$. 

\begin{algorithm*}[!htbp]
\DontPrintSemicolon
\caption{Training \ours{}}
\label{alg:model}
Randomly Initialize $\phi$ network \tcp*[r]{$L2$ normalized output}
Randomly Initialize $\psi$ network \tcp*[r]{$dim(output) = \#A \times dim(W)$}
\For{$e := 1,\infty$}{
    sample $w$ from $L2$ normalized $\mathcal{N}(0,I(dim(W)))$ \tcp*[r]{uniform ball}
    $Q(\cdot, a | w) \leftarrow \psi(\cdot,a, w)^\top w, \forall a \in A$\;
    \For{$t := 1,T$}{
        Receive observation $s_t$ from environment \;
                
            $a_t \leftarrow \epsilon$-greedy policy based on $Q(s_t,\cdot | w)$ \;
        Take action $a_t$, receive observation $s_{t+1}$ and reward \textcolor{orange}{\bcancel{$r_t$}} from environment\;
        $a' = \argmax_a \psi(s_{t+1},a, w)^\top w$\;
        \textcolor{orange}{Compute $r_{\text{\ours{}}}(s_t, a, s_{t+1})$ with~\Eqref{eq:aps_reward}} \tcp*[r]{intrinsic reward to $\max I(s; z)$}
        \textcolor{orange}{$y = r_{\text{\ours{}}}(s_t, a, s_{t+1}) + \gamma \psi(s_{t+1},a', w)^\top w$}\;
        $loss_\psi = (\psi(s_t,a_t, w)^\top w - y_i)^2$\;
        $loss_\phi = -\phi(s_t)^\top w$ \tcp*[r]{minimize Von-Mises NLL}
        Gradient descent step on $\psi$ and $\phi$ \tcp*[r]{minibatch in practice}
    }
}
\end{algorithm*}

This intractable conditional entropy can be lower-bounded by a variational approximation, 
\begin{align*}
    F = - H(s|z) \geq \E_{s, z} \left[ \log q(s|z) \right].
\end{align*}
This is because of the variational lower bound~\citep{Barber2003TheIA}.
\begin{align}
    F & = \sum_{s, z} p(s,z) \log p(s|z) \nonumber \\
    &= \sum_{s,z} p(s,z) \log p(s|z) + \sum_{s,z} p(s,z) \log q(s|z)  \nonumber  \\
    & - \sum_{s,z} p(s,z) \log q(s|z) \nonumber  \\
    &= \sum_{s,z} p(s,z) \log q(s|z) + \sum_z p(z)\KL(p(\cdot|z)||q(\cdot|z)) \nonumber  \\
    &\geq \sum_{s,z} p(s,z) \log q(s|z) \nonumber \\
    &= \mathbb{E}_{s, z} [\log q(s|z)]
\end{align}

Our key observation is that Von Mises-Fisher distribution is symmetric to its parametrization, by restricting $z \equiv w$ similarly to VISR, the reward can be written as
\begin{align}
    \label{eq:aps_exploitation}
    r_{\text{APS}}^{\text{exploitation}}(s, a, s') = \log q(s|w) = \phi(s)^T w.
\end{align}
We find it helps training by sharing the weights between encoders $f$ and $\phi$. 
The encoder is trained by minimizing the negative log likelihood of Von-Mises distribution $q(s|w)$ over the data.
\begin{align}
    L = - \E_{s, w}\left[\log q(s|w)\right] = - \E_{s, w}\left[ \phi(s_t)^\top w\right].
\end{align}

Note that the proposed method is independent from the choices of representation learning for $f$, $e.g.$, one can use an inverse dynamic model~\citep{pathak2017curiosity, burda2018large} to learn the neural encoder, which we leave for future work.

Put~\Eqref{eq:aps_exploration} and~\Eqref{eq:aps_exploitation} together, our intrinsic reward can be written as 
\begin{align}
    & r_{\text{APS}}(s, a, s') \nonumber \\
    & \quad = r_{\text{APS}}^{\text{exploitation}}(s, a, s') + r_{\text{APS}}^{\text{exploration}}(s, a, s') \\ 
    & \quad = \phi(s)^T w + \log \left(1 + \frac{1}{k} \sum_{h^{(j)} \in \mathrm{N}_k(h)} \|h-h^{(j)}\|_{n_h}^{n_h} \right) \label{eq:aps_reward} \nonumber \\
    & \quad\text{where}~ h =\phi(s'),
\end{align}
The output layer of $\phi$ is L2 normalized, task vector $w$ is randomly sampled from a uniform distribution over the unit circle. 

\Tabref{tab:model_summary} positions our new approach with respect to existing ones. 
\Figref{fig:model_diagram} shows the resulting model. Training proceeds as in other algorithms maximizing mutual information: by randomly sampling a task vector $w$ and then trying to infer the state produced by the conditioned policy from the task vector.
\Algref{alg:model} shows the pseudo-code of \ours{}, we highlight the changes from VISR to \ours{} in color.

\subsection{Implementation Details}
We largely follow~\citet{Hansen2020Fast} for hyperparameters used in our Atari experiments, with the following three exceptions. 
We use the four layers convolutional network from~\citet{kostrikov2020image} as the encoder $\phi$ and $f$.
We change the output dimension of the encoder from 50 to 5 in order to match the dimension used in VISR.
While VISR incorporated LSTM~\citep{hochreiter1997long} we excluded it for simplicity and accelerating research.
We use ELU nonlinearities~\citep{clevert2015fast} in between convolutional layers.
We do not use the distributed training setup in~\citet{Hansen2020Fast}, after every roll-out of 10 steps, the experiences are added to a replay buffer.
This replay buffer is used to calculate all of the losses and change the weights of the network.
The task vector $w$ is also resampled every 10 steps. We use n-step Q-learning with $n=10$.

Following~\citet{Hansen2020Fast}, we condition successor features on task vector, making $\psi(s, a, w)$ a UVFA~\citep{borsa2018universal, schaul2015universal}.
We use the Adam optimizer~\citep{kingma2014adam} with an learning rate $0.0001$.
We use discount factor $\gamma = .99$. 
Standard batch size of 32. 
$\psi$ is coupled with a target network~\citep{mnih2015human}, with an update period of $100$ updates.

\section{Results}

\begin{table*}[t]
    \caption{Performance of different methods on the 26 Atari games considered by~\citep{kaiser2019model} after 100K environment steps. The results are recorded at the end of training and averaged over 5 random seeds for \ours{}. \ours{} outperforms prior methods on all aggregate metrics, and exceeds expert human performance on 8 out of 26 games while using a similar amount of experience.}
    \label{tab:atari26_result}
\begin{small}
\centering
\begin{center}
\scalebox{.95}{
\centering
\begin{tabular}{lllllllllll}
\hline
Game                & Random  & Human   & SimPLe          & DER     & CURL    & DrQ     & SPR              & VISR             & APT             & APS (ours)       \\ \hline
Alien               & 227.8   & 7127.7  & 616,9           & 739.9   & 558.2   & 771.2   & 801.5            & 364.4            & \textbf{2614.8} & 934.9            \\
Amidar              & 5.8     & 1719.5  & 88.0            & 188.6   & 142.1   & 102.8   & 176.3            & 186.0            & \textbf{211.5}  & 178.4            \\
Assault             & 222.4   & 742.0   & 527.2           & 431.2   & 600.6   & 452.4   & 571.0            & \textbf{12091.1} & 891.5           & 413.3            \\
Asterix             & 210.0   & 8503.3  & 1128.3          & 470.8   & 734.5   & 603.5   & 977.8            & \textbf{6216.7}  & 185.5           & 1159.7           \\
Bank Heist          & 14.2    & 753.1   & 34.2            & 51.0    & 131.6   & 168.9   & 380.9            & 71.3             & \textbf{416.7}  & 262.7            \\
BattleZone          & 2360.0  & 37187.5 & 5184.4          & 10124.6 & 14870.0 & 12954.0 & 16651.0          & 7072.7           & 7065.1          & \textbf{26920.1} \\
Boxing              & 0.1     & 12.1    & 9.1             & 0.2     & 1.2     & 6.0     & 35.8             & 13.4             & 21.3            & \textbf{36.3}    \\
Breakout            & 1.7     & 30.5    & 16.4            & 1.9     & 4.9     & 16.1    & 17.1             & 17.9             & 10.9            & \textbf{19.1}    \\
ChopperCommand      & 811.0   & 7387.8  & 1246.9          & 861.8   & 1058.5  & 780.3   & 974.8            & 800.8            & 317.0           & \textbf{2517.0}  \\
Crazy Climber       & 10780.5 & 23829.4 & 62583.6         & 16185.2 & 12146.5 & 20516.5 & 42923.6          & 49373.9          & 44128.0         & \textbf{67328.1} \\
Demon Attack        & 107805  & 35829.4 & 62583.6         & 16185.3 & 12146.5 & 20516.5 & 42923.6          & \textbf{8994.9}  & 5071.8          & 7989.0           \\
Freeway             & 0.0     & 29.6    & 20.3            & 27.9    & 26.7    & 9.8     & 24.4             & -12.1            & \textbf{29.9}   & 27.1             \\
Frostbite           & 65.2    & 4334.7  & 254.7           & 866.8   & 1181.3  & 331.1   & \textbf{1821.5}  & 230.9            & 1796.1          & 496.5            \\
Gopher              & 257.6   & 2412.5  & 771.0           & 349.5   & 669.3   & 636.3   & 715.2            & 498.6            & \textbf{2590.4} & 2386.5           \\
Hero                & 1027.0  & 30826.4 & 2656.6          & 6857.0  & 6279.3  & 3736.3  & 7019.2           & 663.5            & 6789.1          & \textbf{12189.3} \\
Jamesbond           & 29.0    & 302.8   & 125.3           & 301.6   & 471.0   & 236.0   & 365.4            & 484.4            & 356.1           & \textbf{622.3}   \\
Kangaroo            & 52.0    & 3035.0  & 323.1           & 779.3   & 872.5   & 940.6   & 3276.4           & 1761.9           & 412.0           & \textbf{5280.1}  \\
Krull               & 1598.0  & 2665.5  & \textbf{4539.9} & 2851.5  & 4229.6  & 4018.1  & 2688.9           & 3142.5           & 2312.0          & 4496.0           \\
Kung Fu Master      & 258.5   & 22736.3 & 17257.2         & 14346.1 & 14307.8 & 9111.0  & 13192.7          & 16754.9          & 17357.0         & \textbf{22412.0} \\
Ms Pacman           & 307.3   & 6951.6  & 1480.0          & 1204.1  & 1465.5  & 960.5   & 1313.2           & 558.5            & \textbf{2827.1} & 2092.3           \\
Pong                & -20.7   & 14.6    & \textbf{12.8}   & -19.3   & -16.5   & -8.5    & -5.9             & -26.2            & -8.0            & 12.5             \\
Private Eye         & 24.9    & 69571.3 & 58.3            & 97.8    & 218.4   & -13.6   & \textbf{124.0}   & 98.3             & 96.1            & 117.9            \\
Qbert               & 163.9   & 13455.0 & 1288.8          & 1152.9  & 1042.4  & 854.4   & 669.1            & 666.3            & 17671.2         & \textbf{19271.4} \\
Road Runner         & 11.5    & 7845.0  & 5640.6          & 9600.0  & 5661.0  & 8895.1  & \textbf{14220.5} & 6146.7           & 4782.1          & 5919.0           \\
Seaquest            & 68.4    & 42054.7 & 683.3           & 354.1   & 384.5   & 301.2   & 583.1            & 706.6            & 2116.7          & \textbf{4209.7}  \\
Up N Down           & 533.4   & 11693.2 & 3350.3          & 2877.4  & 2955.2  & 3180.8  & \textbf{28138.5} & 10037.6          & 8289.4          & 4911.9           \\ \hline
Mean Human-Norm'd   & 0.000   & 1.000   & 44.3            & 28.5    & 38.1    & 35.7    & 70.4             & 64.31            & 69.55           & \textbf{99.04}   \\
Median Human-Norm'd & 0.000   & 1.000   & 14.4            & 16.1    & 17.5    & 26.8    & 41.5             & 12.36            & 47.50           & \textbf{58.80}   \\ \hline
\# Superhuman       & 0       & N/A     & 2               & 2       & 2       & 2       & 7                & 6                & 7               & \textbf{8}       \\ \hline
\end{tabular}
}
\end{center}
\end{small}
\vspace{-1em}
\end{table*}

We test \ours{} on the full suite of 57 Atari games~\citep{bellemare2013arcade} and the sample-efficient Atari setting~\citep{kaiser2019model, van2019use} which consists of the 26 easiest games in the Atari suite (as judged by above random performance for their algorithm).

We follow the evaluation setting in VISR~\citep{Hansen2020Fast} and APT~\citep{liu2021behavior}, agents are allowed a long unsupervised training phase (250M steps) without access to rewards, followed by a short test phase with rewards. 
The test phase contains 100K environment steps – equivalent to 400k frames, or just under two hours – compared to the typical standard of 500M environment steps, or roughly 39 days of experience.
We normalize the episodic return with 
respect to expert human scores to account for different scales of scores in each game, as done in previous works. 
The human-normalized performance of an agent on a game is calculated as $\frac{\text{agent score} - \text{random score}}{\text{human score} - \text{random score}}$ and aggregated across games by mean or median.

When testing the pre-trained successor features $\psi$, we need to find task vector $w$ from the rewards. To do so, we rollout 10 episodes (or 40K steps, whichever comes ﬁrst) with the trained \ours{}, each conditioned on a task vector chosen uniformly on a 5-dimensional sphere. From these initial episodes, we combine the data across all episodes and solve the linear regression problem. 
Then we fine-tune the pre-trained model for 60K steps with the inferred task vector, and the average returns are compared.

A full list of scores and aggregate metrics on the Atari 26 subset is presented in~\Tabref{tab:atari26_result}.
The results on the full 57 Atari games suite is presented in Supplementary Material. 
For consistency with previous works, we report human and random scores from~\citep{hessel2018rainbow}.

In the data-limited setting, \ours{} achieves super-human performance on eight games and achieves scores higher than previous state-of-the-arts. 

In the full suite setting, \ours{} achieves super-human performance on 15 games, compared to a maximum of 12 for any previous methods and achieves scores significantly higher than any previous methods.

\section{Analysis}

\paragraph{Contribution of Exploration and Exploitation}

\begin{figure}[!htbp]
\centering
\includegraphics[width=.48\textwidth]{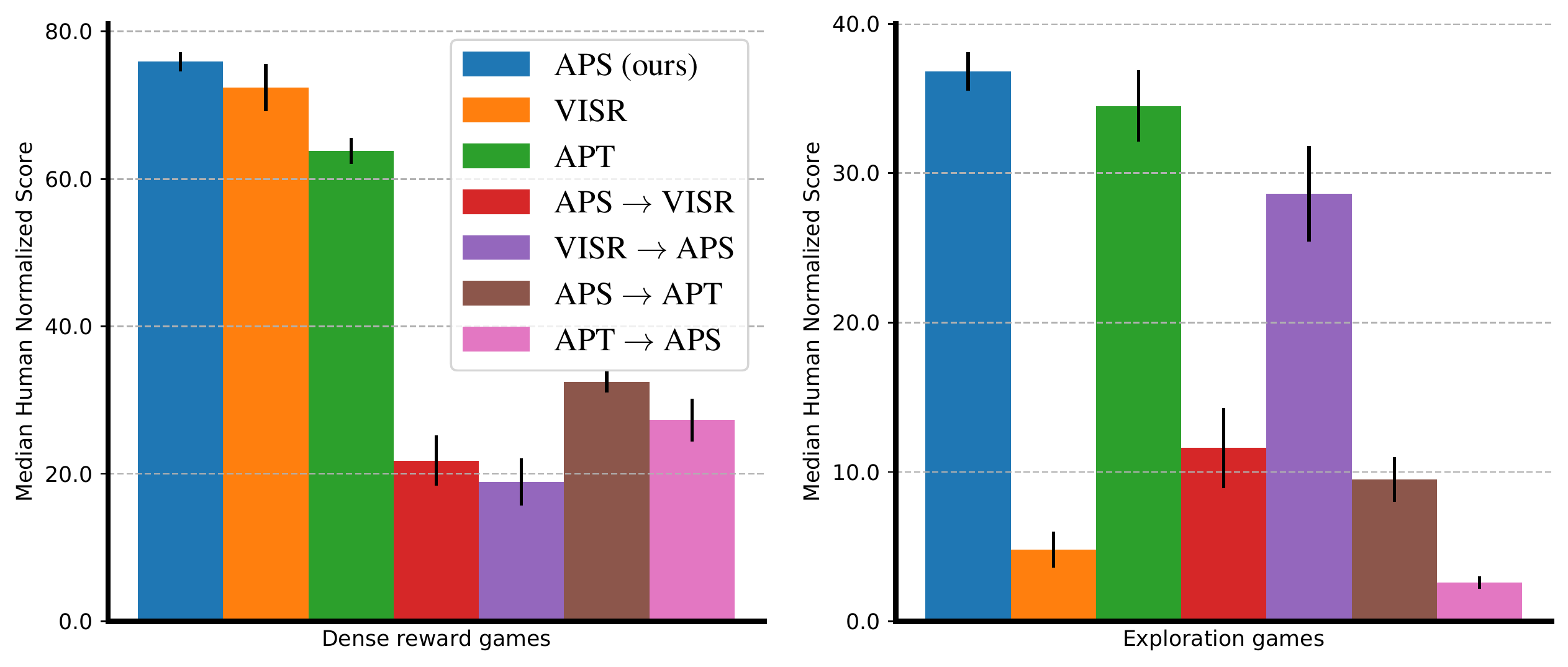}
\caption{Scores of different methods and their variants on the 26 Atari games considered by~\citet{kaiser2019model}. $X \rightarrow Y$ denotes training method $Y$ using the data collected by method $X$ at the same time. 
}
\label{fig:atari_ablate_result}
\end{figure}

In order to measure the contributions of components in our method, we aim to answer the following two questions in this ablation study.
Compared with APT ($\max H(s)$), is the improvement solely coming from better fast task solving induced by $\max -H(s|z)$ and the exploration is the same?
Compared with VISR ($\max H(z) - H(z|s)$), is the improvement solely coming from better exploration due to $\max H(s) - H(s|z)$ and the task solving ability is the same?

We separate Atari 26 subset into two categories. Dense reward games in which exploration is simple and exploration games which require exploration.
In addition to train the model as before, we simultaneously train another model using the same data, $e.g.$ $\text{\ours} \rightarrow \text{APT}$ denotes when training \ours{} simultaneously training APT using the same data as \ours{}. 
As shown in~\Figref{fig:atari_ablate_result}, on dense reward games, $\text{\ours} \rightarrow \text{APT}$ performs better than $\text{APT} \rightarrow \text{\ours}$.
On exploration games, $\text{\ours} \rightarrow \text{APT}$ significantly outperforms $\text{APT} \rightarrow \text{\ours}$.
Similarly $\text{\ours} \rightarrow \text{VISR}$ performs better than the other way around.
Together, the results indicate that entropy maximization and variational successor features improves each other in a nontrivial way, and both are important to the performance gain of \ours{}.

\begin{table}[!htbp]
    \caption{Scores on the 26 Atari games for variants of \ours{}, VISR, and APT. Scores of considered variants are averaged over 3 random seeds.}
    \label{tab:fine_tune_ablation}
    \centering
    \scalebox{0.85}{
    \begin{tabular}{l r r}
    \toprule
    Variant & \multicolumn{2}{c}{Human-Normalized Score}  \\
    {} & mean & median \\
    \midrule
    \ours{} &    \textbf{99.04}  & \textbf{58.80} \\
    \midrule
    \ours{} w/o fine-tune & 81.41  & 49.18  \\
    VISR (controlled, w/ fine-tune) & 68.95 & 31.87 \\
    APT (controlled, w/o fine-tune) & 58.23 & 19.85 \\
    \ours{} w/o shared encoder & 87.59 & 51.45 \\
    \bottomrule
    \end{tabular}
    }
\end{table}
\vspace{-1em}

\paragraph{Fine-Tuning Helps Improve Performance}

We remove fine-tuning from \ours{} that is we evaluate its zero-shot performance, the same as in~\citet{Hansen2020Fast}.
We also employ \ours{}'s fine-tuning scheme to VISR, namely 250M (without access to rewards, followed by a short task identify phase (40K steps) and a fine-tune phase (60K steps).
The results shown in~\Tabref{tab:fine_tune_ablation} demonstrate that fine-tuning can boost performance. 
\ours{} w/o fine-tune outperforms all controlled baselines, including VISR w/ fine-tune.

\paragraph{Shared Encoder Can Boost Data-Efficiency}

We investigate the effect of using $\phi$ as the encoder $f$. 
To do so, we consider a variant of \ours{} that learns the encoder $f$ as in APT by contrastive representation learning. 
The performance of this variant is denoted as \ours{} w/o shared encoder shown in~\Tabref{tab:fine_tune_ablation}. 
Sharing encoder can boost data efficiency, we attribute the effectiveness to $\phi$ better captures the relevant information which is helpful for computing intrinsic reward.
We leave the investigation of using other representation learning methods as future work.

\section{Conclusion}
In this paper, we propose a new unsupervised pretraining method for RL. 
It addresses the limitations of prior mutual information maximization-based and entropy maximization-based methods and combines the best of both worlds.
Empirically, \ours{} achieves state-of-the-art performance on the Atari benchmark, demonstrating signiﬁcant improvements over prior work.

Our work demonstrates the benefit of leveraging state entropy maximization data for task-conditioned skill discovery.
We are excited about the improved performance by decomposing mutual information as $H(s) - H(s|z)$ and optimizing them by particle-based entropy and variational successor features. 
In the future, it is worth studying how to combine approaches designed for maximizing the alternative direction $- H(z|s)$ with the particle-based entropy maximization.
\section{Acknowledgment}
We thank members of Berkeley Artificial Intelligence Research (BAIR) Lab for many insightful discussions.
This work was supported by Berkeley Deep Drive, the Open Philanthropy Project, and Intel. 

\bibliography{refs.bib}
\bibliographystyle{icml2021}

\clearpage
\appendix
\section{Experiment Details}

The corresponding hyperparameters used in Atari experiments are shown in~\Tabref{tab:hyper_params_atari} and~\Tabref{tab:hyper_params_atari_reward}.
We follow~\citet{kostrikov2020image} to use data augmentation techniques that consist of a simple random shift which has been shown effective in visual domain RL. 
Specifically, the images are padded each side by 4 pixels (by repeating boundary pixels) and then select a random $84 \times 84$ crop, yielding the original image. 
This procedure is repeated every time an image is sampled from the replay buffer.

We also use the same generalized policy improvement(GPI)~\citep{barreto2017successor, barreto2020fast} as in VISR with the number of polices 10. GPI is also used in VISR to ensure a fair comparison.
Per common practice, we average performance of our agent over 5 random seeds. The evaluation is done for 125K environment steps at the end of training for 100K environment steps.
We follow~\citet{Hansen2020Fast} to use one MLP for each dimension of the successor feature. 

The ablated variant \ours{} w/o shared encoder follows APT~\citep{liu2021behavior} but the output dimension of the neural encoder $f$ is decreased to 5 in order to match the default \ours{}.
The projection network in contrastive learning is a two-layer MLP with hidden size of 128 and output size of 64.
We also use the same temperature and other hyperparamters as APT for the ablation study.

\begin{table*}[!htbp]
\caption{Hyper-parameters for RL.}
\label{tab:hyper_params_atari}
\centering
\begin{tabular}{|l|c|}
\hline
Parameter         & Setting \\
\hline
Terminal on loss of life & True \\
Reward clipping (fine-tuning phase) & $[-1, 1]$ \\
Data augmentation & Random shifts and Intensity \\
Grey-scaling & True \\
Observation down-sampling & $84 \times 84$ \\
Frames stacked & $4$ \\
Action repetitions & $4$ \\
Max frames per episode & $108$k \\
Update  &  Double Q \\
Target network: update period & $100$ \\
Discount factor &  $0.99$ \\
Minibatch size  & $32$ \\
$\psi, \phi$ optimizer  & Adam \\
$\psi, \phi$ optimizer (pre-training phase): learning rate & $0.0001$\\
$\psi, \phi$ optimizer (fine-tuning phase): learning rate & $0.001$\\
$\psi, \phi$ optimizer: $\beta_1$  & $0.9$ \\
$\psi, \phi$ optimizer: $\beta_2$ &  $0.999$ \\
$\psi, \phi$ optimizer: $\epsilon$ &  $0.00015$ \\
Max gradient norm & $10$ \\
Training steps (fine-tuning phase) &  60K \\
Task identity steps (fine-tuning phase) & 40K \\
Training steps (pre-training phase) & 5M \\
Evaluation steps & 125K \\
Min replay size for sampling& $1600$\\
Memory size & Unbounded \\
Replay period every & $1$ step \\
Multi-step return length & $10$ \\
$\psi$ network: channels  & $32, 64, 64$ \\ 
$\psi$ network: filter size  & $8 \times 8$, $4 \times 4$, $3 \times 3$ \\
$\psi$ network: stride  & $4, 2, 1$ \\
$\psi$ network: hidden units  & $512$ \\
$\psi$ Non-linearity & \texttt{ReLU}\\

Exploration & $\epsilon$-greedy \\
$\epsilon$-decay & $2500$ \\
\hline
\end{tabular}\\
\end{table*}

\begin{table*}[!htbp]
\caption{Hyper-parameters for Learning $\phi$.}
\label{tab:hyper_params_atari_reward}
\centering
\begin{tabular}{|l|c|}
\hline
Parameter         & Setting \\
\hline
Value of k & search in $\{3, 5, 10\}$ \\
$\phi$ network: channels  & $32, 64, 64$ \\ 
$\phi$ network: filter size  & $8 \times 8$, $4 \times 4$, $3 \times 3$ \\
$\phi$ network: stride  & $4, 2, 1$ \\
$\phi$ network: hidden units  & $512$ \\
$\phi$ network Non-linearity & \texttt{ELU}\\
FC hidden size & 1024 \\ 
Output size & 5 \\
\hline
\end{tabular}\\
\end{table*}

\section{Scores Breakdown on 57 Atari games}
A comparison between \ours{} and baselines on each individual game of the 57 Atari game suite is shown in~\Tabref{tab:atari57_result}.
\ours{} achieves super-human performance on 15 games, compared to a maximum of 12 for any previous methods and achieves scores significantly higher than any previous methods.

\begin{table*}[!htbp]
\centering
\caption{Comparison of raw scores of each method on Atari games. Results are averaged over five random seeds. $@N$ represents the amount of RL interaction utilized at fine-tuning phase.
}
\label{tab:atari57_result}
\setlength{\tabcolsep}{6.0pt}
\renewcommand{\arraystretch}{0.85}
\centering
\begin{center}
\scalebox{.95}{
\centering
\begin{tabular}{l|ll|lll}
\hline
Game                & Random   & Human   & VISR              & APT             & \ours{} (ours)                           \\ \hline
Alien               & 227.8    & 7127.7  & 364.4             & \textbf{2614.8} & 934.9                         \\
Amidar              & 5.8      & 1719.5  & 186.0             & \textbf{211.5}  & 188.4                         \\
Assault             & 222.4    & 742.0   & \textbf{1209.1}   & 891.5           & 413.3                         \\
Asterix             & 210.0    & 8503.3  & \textbf{6216.7}   & 185.5           & 1159.5                        \\
Asteroids           & 7191     & 47388.7 & \textbf{4443.3}   & 678.7           & {\color[HTML]{333333} 1519.7} \\
Atlantis            & 12850.0  & 29028.1 & 140542.8          & 40231.0         & \textbf{18920.0}              \\
Bank Heist          & 14.2     & 753.1   & 71.3              & \textbf{416.7}  & 262.7                         \\
Battle Zone         & 2360.0   & 37187.5 & 7072.7            & 7065.1          & \textbf{26920.1}              \\
Beam Rider          & 363.9    & 16826.5 & 1741.9            & 3487.2          & \textbf{4981.2}               \\
Berzerk             & 123.7    & 2630.4  & 490.0             & \textbf{493.4}  & 387.4                         \\
Bowling             & 23.1     & 160.7   & 21.2              & -56.5           & \textbf{56.5}                 \\
Boxing              & 0.1      & 12.1    & 13.4     & 21.3            & \textbf{36.3}                         \\
Breakout            & 1.7      & 30.5    & 17.9              & 10.9            & \textbf{19.1}                 \\
Centipede           & 2090.9   & 12017.1 & \textbf{7184.9}   & 6233.9          & 3915.7                        \\
Chopper Command     & 811.0    & 7387.8  & 800.8             & 317.0           & \textbf{2517.0}               \\
Crazy Climber       & 10780.5  & 23829.4 & 49373.9           & 44128.0         & \textbf{67328.1}              \\
Defender            & 2874.5   & 18688.9 & 15876.1           & 5927.9          & \textbf{19921.5}              \\
Demon Attack        & 107805   & 35829.4 & \textbf{8994.9}   & 6871.8          & 7989.0                        \\
Double Dunk         & -18.6    & -16.4   & -22.6             & -17.2           & \textbf{-8.0}                 \\
Enduro              & 0.0      & 860.5   & -3.1              & -0.3            & \textbf{216.8}                \\
Fishing Derby       & -91.7    & -38.7   & -93.9             & -5.6            & \textbf{-2.1}                 \\
Freeway             & 0.0      & 29.6    & -12.1             & \textbf{29.9}   & 27.1                          \\
Frostbite           & 65.2     & 4334.7  & 230.9             & \textbf{1796.1} & 496.1                         \\
Gopher              & 257.6    & 2412.5  & 498.6             & \textbf{2190.4} & 2590.4                        \\
Gravitar            & 173.0    & 3351.4  & 328.1             & \textbf{542.0}  & 487.0                         \\
Hero                & 1027.0   & 30826.4 & 663.5             & 6789.1          & \textbf{12189.3}              \\
Ice Hockey          & -11.2    & 0.9     & -18.1             & -30.1           & \textbf{-11.3}                \\
Jamesbond           & 29.0     & 302.8   & 484.4             & 356.1           & \textbf{622.3}                \\
Kangaroo            & 52.0     & 3035.0  & 1761.9            & 412.0           & \textbf{5280.1}               \\
Krull               & 1598.0   & 2665.5  & 3142.5            & 2312.0          & \textbf{4496.0}               \\
Kung Fu Master      & 258.5    & 22736.3 & 16754.9           & 17357.0         & 13112.1                       \\
Montezuma Revenge   & 0.0      & 4753.3  & 0.0               & 147.0           & \textbf{211.0}                \\
Ms Pacman           & 307.3    & 6951.6  & 558.5             & \textbf{2527.1} & 2092.3                        \\
Name This Game      & 2292.3   & 8049.0  & 2605.8            & 1387.2          & \textbf{6898.8}               \\
Phoenix             & 761.4    & 7242.6  & \textbf{7162.2}   & 3874.2          & 6871.8                        \\
Pitfall             & -229.4   & 6463.7  & -370.8            & -12.8           & \textbf{-6.2}                 \\
Pong                & -20.7    & 14.6    & -26.2             & -8.0            & \textbf{12.5}                 \\
Private Eye         & 24.9     & 69571.3 & 98.3              & 96.1            & \textbf{117.9}                \\
Qbert               & 163.9    & 13455.0 & 666.3             & 17671.2         & \textbf{19271.4}              \\
Riverraid           & 1338.5   & 17118.0 & 5422.2            & 4671.0          & \textbf{10521.3}              \\
Road Runner         & 11.5     & 7845.0  & \textbf{6146.7}   & 4782.1          & 5919.0                        \\
Robotank            & 2.2      & 11.9    & 10.0              & \textbf{13.7}   & 12.6                          \\
Seaquest            & 68.4     & 42054.7 & 706.6             & 2116.7          & \textbf{4209.7}               \\
Skiing              & -17098.1 & -4336.9 & \textbf{-19692.5} & -38434.1        & -9102.1                       \\
Solaris             & 1236.3   & 12326.7 & \textbf{1921.5}   & 841.8           & 1095.4                        \\
Space Invaders      & 148.0    & 1668.7  & \textbf{9741.0}   & 3687.2          & 3693.8                        \\
Star Gunner         & 664.0    & 10250.0 & 25827.5           & 8717.0          & \textbf{42970.0}              \\
Surround            & -10.0    & 6.5     & -15.5             & \textbf{-2.5}   & -5.8                          \\
Tennis              & -23.8    & -8.3    & 0.7               & 1.2             & \textbf{8.7}                  \\
Time Pilot          & 3568.0   & 5229.2  & 4503.6            & 2567.0          & \textbf{4586.5}               \\
Tutankham           & 11.4     & 167.6   & 50.7              & \textbf{124.6}  & 45.6                          \\
Up N Down           & 533.4    & 11693.2 & \textbf{10037.6}  & 8289.4          & 4911.9                        \\
Venture             & 0.0      & 1187.5  & -1.7              & \textbf{231.0}  & 136.0                         \\
Video Pinball       & 0.0      & 17667.9 & \textbf{35120.3}  & 2817.1          & 154414.1                      \\
Wizard Of Wor       & 563.5    & 4756.5  & 853.3             & 1265.0          & \textbf{1732.1}               \\
Yars Revenge        & 3092.9   & 54576.9 & 5543.5            & 1871.5          & \textbf{6539.5}               \\
Zaxxon              & 32.5     & 9173.3  & 897.5             & 3231.0          & \textbf{5819.2}               \\ \hline
Mean Human-Norm'd   & 0.000    & 1.000   & 68.42             & 47.78           & \textbf{103.04}               \\
Median Human-Norm'd & 0.000    & 1.000   & 9.41              & 33.41           & \textbf{39.23}                \\ \hline
\#Superhuman        & 0        & N/A     & 11                & 12              & \textbf{15}                   \\ \hline
\end{tabular}
}
\end{center}
\end{table*}

\end{document}